\documentclass{article}
\usepackage{ijcai22}

\usepackage{times}
\usepackage{soul}
\usepackage{url}
\usepackage[hidelinks]{hyperref}
\usepackage[utf8]{inputenc}
\usepackage[small]{caption}
\usepackage{graphicx}
\usepackage{amsmath}
\usepackage{amsthm}
\usepackage{booktabs}
\usepackage{algorithm}
\usepackage{algorithmic}
\title{Self-Evolutionary Clustering}

\author{
Hanxuan Wang
\and
Na Lu$^*$\and
Qinyang Liu
\affiliations
School of Automation Science and Engineering, Xi'an Jiaotong University
\emails
wanghanxuan1995@stu.xjtu.edu.cn,
lvna2009@xjtu.edu.cn,
qinyang@stu.xjtu.edu.cn
}

\begin{document}

\maketitle

\begin{abstract}
Deep clustering outperforms conventional clustering by mutually promoting representation learning and cluster assignment. However, most existing deep clustering methods suffer from two major drawbacks. First, most cluster assignment methods are based on simple distance comparison and highly dependent on the target distribution generated by a handcrafted nonlinear mapping. These facts largely limit the possible performance that deep clustering methods can reach. Second, the clustering results can be easily guided towards wrong direction by the misassigned samples in each cluster. The existing deep clustering methods are incapable of discriminating such samples. To address these issues, a novel modular Self-Evolutionary Clustering (Self-EvoC) framework is constructed, which boosts the clustering performance by classification in a self-supervised manner. Fuzzy theory is used to score the sample membership with probability which evaluates the intermediate clustering result certainty of each sample. Based on which, the most reliable samples can be selected and augmented. The augmented data are employed to fine-tune an off-the-shelf deep network classifier with the labels from the clustering, which results in a model to generate the target distribution. The proposed framework can efficiently discriminate sample outliers and generate better target distribution with the assistance of self-supervised classifier. Extensive experiments indicate that the Self-EvoC remarkably outperforms state-of-the-art deep clustering methods on three benchmark datasets.
\end{abstract}

\section{Introduction}

In comparison to classification, clustering aims at discovering patterns within data in an unsupervised manner. In practical application, collection of labeled data might be very difficult or quite expensive, e.g. medical images, forest fire images, fault data of aircraft engine and so on. In some occasions, there might be unseen categories which also lack supervision information. Therefore, even though classification methods based on deep learning have witnessed rapid development, the research on deep clustering is far behind \cite{xie2016unsupervised,guo2017improved}. Clustering methods have not been able to benefit from the development of classification so far.

The conventional clustering methods such as $k$-means \cite{macqueen1967some}, spectral clustering \cite{von2007tutorial}, DBSCAN \cite{ester1996density} map data into different clusters according to the distance evaluated in the original data space or handcrafted feature space. For these methods, how to reduce the feature dimension and preserve the local structure buried in the data are of essential meaning. The representative dimension reduction methods include linear method PCA, nonlinear dimension reduction by locally linear embedding (LLE) \cite{roweis2000nonlinear} and so on. The feature learning ability of these algorithms is rather limited as compared with the fast developed deep learning models \cite{lu2021transferable}. With the help of DNN feature extractor, clustering has shown better performance, which is specially called Deep Clustering (DC) \cite{xie2016unsupervised,peng2016deep,lin2021graph}.
Most deep clustering algorithms are trained by mutually promoting representation learning and cluster assignment according to the difference between the cluster assignment and target distribution \cite{li2019deep}. These methods have achieved some improvement by learning clustering toward features. However, there are still some problems unsolved. In this paper, we focus on two problems.

First, the target distribution in deep clustering model highly depends on artificially designed distribution function \cite{xie2016unsupervised,guo2017improved}. How can we automatically generate the target distribution suitable for specific dataset remains a problem. Deep clustering algorithms mainly have two stages, feature transformation and cluster assignment. The two stages promote each other based on clustering loss. The clustering loss evaluates the difference between the obtained intermediate data distribution and target distribution computed by an artificially designed function with the obtained intermediate distribution as input. Therefore, the target distribution generating function restricts the best performance of existing deep clustering models.

Second, misassginments of samples are inevitable in clustering and can lead clustering towards wrong direction. How to select the confidential samples for each cluster and reduce the influence from the possible misassignment to guide the clustering toward correct direction is another unsolved problem. Some researches have tried to solve the problem with self-paced learning \cite{guo2019adaptive,zhou2020self}. These solutions evaluate the easiness of samples based on the distance of the sample to the cluster centers. However, the cluster centers are usually not uniformly distributed which will bias the easiness calculation of samples belonging to different clusters. Meanwhile, these solutions incorporate extra parameters like the age parameter controlling the learning pace and step size to change the age parameter. These parameters need be set manually. 

To address the above two problems, Self-EvoC framework is developed combining three modules including feature extractor, clustering and self-supervised classifier. Self-EvoC works in a modular manner where the three modules can be replaced by any model toward application. In our solution, different feature extractors are adopted including Autoencoder and off-the-shelf deep models. Fuzzy clustering is used as the clustering module which can naturally gives the confidence level of samples. Based on which, the confidential samples can be selected and the controversial samples can be spotted. The selected confidential samples are augmented with techniques like random rotating and cropping etc. The augmented data are fed to the following classifier to train a target distribution generation model in a self-supervised manner with the clustering labels. Off-the-shelf network model is employed as the classifier where no supervision is needed. The classifier is fine-tuned by the reliable augmented data and then used to generate the target distribution. In comparison to the manually defined target distribution generating function in existing deep clustering, the target distribution generating block becomes a trainable neural network module which is more reliable and adaptive. In this way, the clustering model get evolved with the motivation from the classifier in a self-supervised manner. The performance of Self-EvoC has been evaluated on three benchmarks and compared with multiple state-of-the-art methods. The experiment results verified the effectiveness of Self-EvoC.
The contributions are summarized as follows.
\begin{itemize}
	\item We propose a novel modular deep clustering framework combining feature extractor, clustering and classifier. With this framework, the clustering model get evolved with the assistance of the neural network classifier in a self-supervised manner. 
	\item Fuzzy theory is introduced into clustering to model the sample fuzzy membership which provides a natural measure for sample confidence. Based on which, the reliable samples and controversial samples can be differentiated. 
	\item Off-the-shelf network model is adopted as the classifier to train a target distribution generation model instead of the manually defined function. The augmented reliable data are used to fine-tune the classifier with the cluster labels self-supervisedly. 
\end{itemize}

\section{Related Work}

Deep Clustering is a family of clustering algorithms which take advantage of the feature extracting ability of DNN to complement the traditional clustering methods. The basic framework of deep clustering is illustrated in Figure 1. Low dimensional features can be obtained by a feature extractor. A reconstruction loss ${L_{\rm{R}}}$ is used to measure the quality of the obtained features for most feature extractors used in deep clustering like Autoencoder. Meanwhile, the target distribution is introduced to optimize the cluster assignment in a self-training way. The target distribution is generated by a manually designed nonlinear function based on the sample assignment result. The difference between the sample assignment and the target distribution is evaluated by a clustering loss denoted as ${L_{\rm{C}}}$. According to the properties of the adopted DNN, deep clustering can be divided into Generative DC and non-Generative DC. Generative DC is mainly based on generative models such as VAE \cite{kingma2013auto} and GAN \cite{wang2018partial}, while non-Generative DC adopts Autoencoder or pre-trained models as feature extractor \cite{xie2016unsupervised,guo2017improved}. For objective function, there are combined-loss \cite{song2014deep} and clustering loss \cite{lin2018deep}. The combined-loss is a combination of reconstruction loss and clustering loss as $L_{\rm{R}} + \gamma L_{\rm{C}}$, where the tradeoff parameter $\gamma$  need be carefully selected. The clustering loss highly depends on the manually generated target distribution.
\begin{figure}
	\begin{center}
		\footnotesize
		\begin{tabular}{c}
			\includegraphics[scale=0.62]{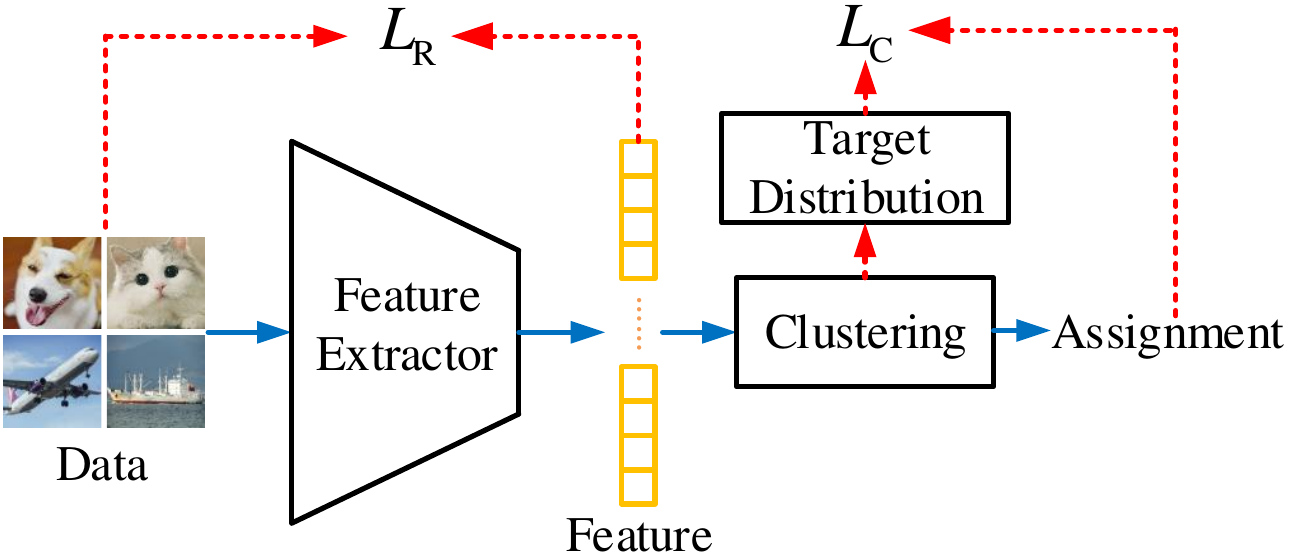}
		\end{tabular}
		\caption{\label{dcmg} {The basic framework of deep clustering.}}
	\end{center}
\end{figure}

Self-EvoC can be categorized as non-Generative clustering loss based clustering. Non-Generative clustering loss based clustering uses ${L_{\rm{C}}}$ to optimize the parameters of the feature extractor and cluster centers. \cite{xie2016unsupervised} proposed a Deep Embedded Clustering (DEC) by defining a KL divergence between the clustering distribution and target distribution as the clustering loss, where the clustering distribution is the soft clustering label and the target distribution is a nonlinear mapping of the cluster distribution. Narrowing the gap between the clustering distribution and target distribution can obtain clustering-friendly features. \cite{li2018discriminatively} proposed a CNN Autoencoder based DEC and modified the normalization method of target distribution. Besides the modification on network structure, \cite{peng2017cascade} designed a clustering loss as the discrepancy between pairwise example-center distributions. \cite{feng2020deep} proposed a clustering model based on graph-regularized deep normalized fuzzy compactness to improve DEC. This method provides flexible cluster labels with fuzzy membership. The target distribution of these methods needs be designed very carefully and can not remove misassignments. In this paper, we try to solve these problems by incorporating the idea of confidential sample selection and self-supervised classification.

\section{Self-Evolutionary Clustering}

In order to obtain better target distribution automatically and constrain clustering direction, Self-EvoC framework is proposed as illustrated in Figure 2. The framework combines three replaceable modules including feature extractor, clustering and classifier, which can be trained in an end-to-end way. Denote a dataset for clustering with $N$ samples as $X = {\left\{ {{x_i} \in {R^D}} \right\}_{i = 1,2, \ldots ,N}}$ where $D$ is the sample dimension. The feature extractor can be any off-the-shelf network model. The features obtained by the feature extractor are assigned to different clusters with fuzzy clustering method which can naturally rate the confidence of the sample membership. According to the fuzzy clustering results, an selected and augmented dataset is generated to train the following classifier. The classifier can be any off-the-shelf deep classifier. The augmented data are constructed by fuzzy selecting strategy, local loundary cleaning and augmentation. Based on the augmented data, the classifier can be further fine-tuned in a self-supervised manner with the intermediate clustering label. Then the classifier is used to predict the target distribution. The clustering loss is measured between the intermediate clustering label and the target distribution, which is used to optimize the network.
\begin{figure}[H]
	\begin{center}
		\footnotesize
		\begin{tabular}{c}
			\includegraphics[scale=0.6]{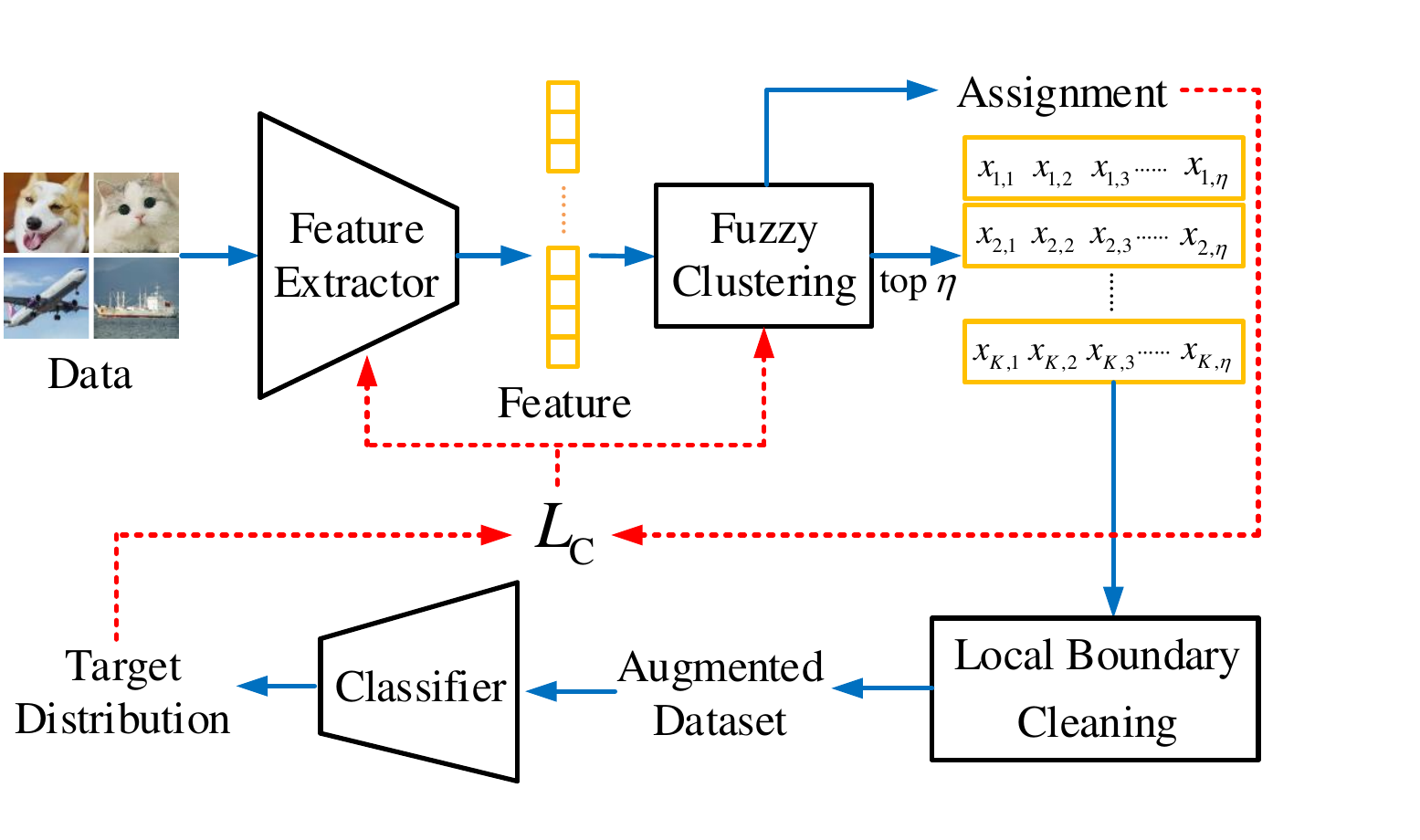}
		\end{tabular}
		\caption{\label{dcmg} { The schema of Self-Evolutionary Clustering.}}
	\end{center}
\end{figure}

\subsection{Fuzzy Clustering based Selecting Strategy}

The major goal of Fuzzy Selecting Strategy is to select high confidence samples for the subsequent classification. A DNN feature extractor $f_w$ is used to transform each example $x_i \in R^D$ to $z_i \in R^L$. Generally speaking, there will be inevitable misassignments after clustering which will influence the clustering performance. At the beginning of the clustering, the features extracted by DNN are not clustering-friendly. As a result, clusters might overlap each other. In general, the closer the sample to the cluster center, the more likely it is to be the right assignment. Based on this hypothesis, most mistakenly assigned samples are supposed to locate close to the cluster boundary. However, different clusters have different distribution shapes, therefore it is inappropriate to select samples only with distance.

In order to solve the above problem, we introduce probability to model the membership confidence. Inspired by FCM (Fuzzy $c$-means) \cite{winkler2011fuzzy}, the membership of the $i^{\rm{th}}$  sample to the $j^{\rm{th}}$ center $u_{ij}$ is formulated as
\begin{align}
{u_{ij}} = \frac{1}{{\sum\limits_{k = 1}^K {{{\left( {\frac{{\left\| {{z_i} - {\mu _j}} \right\|}}{{\left\| {{z_i} - {\mu _k}} \right\|}}} \right)}^{\frac{2}{{m - 1}}}}} }},
\end{align}%
where $m>1$ is the fuzzifier value and $K$ is the number of clusters. This process can be modeled as a clustering layer in the neural network. The corresponding parameters are the clustering center matrix. The objective function to initialize the clustering layer is written as
\begin{align}
	{L_m} = \sum\limits_{i = 1}^N {\sum\limits_{j = 1}^K {u_{ij}^m{{\left\| {{z_i} - {\mu _j}} \right\|}^2}} }, 
\end{align}%
where $N$ is the number of samples. During the process of optimizing Eq. (2), we first calculate the membership with Eq. (1) and then the $j^{\rm{th}}$ clustering center $\mu_j$ is obtained as
\begin{align}
{\mu _j} = {{\sum\limits_{i = 1}^N {u_{ij}^m \cdot {z_i}} } \mathord{\left/
		{\vphantom {{\sum\limits_{i = 1}^N {u_{ij}^m \cdot {z_i}} } {\sum\limits_{i = 1}^N {u_{ij}^m} }}} \right.
		\kern-\nulldelimiterspace} {\sum\limits_{i = 1}^N {u_{ij}^m} }}.
\end{align}%
These two steps are executed alternately until it converges.

Unfortunately, the initialization of the fuzzy clustering layer is unstable. It is observed that the optimization of Eq. (2) can lead to multiple results with the same input and hyper-parameters. To ensure stability, a fuzzy selecting strategy (FSS) is constructed.  After $M$ repeated trials based on the same input and fuzzifier value, we get $M$ membership matrices. The corresponding label of each sample is the clustering center with the largest membership. We treat the result of the first trial as a benchmark and compare it with the remaining $M-1$ predictions. The most frequent result is selected as the most stable initialization of the clustering center $\mu_{\rm{max}}$.

After modeling the sample membership with probability, the membership matrix $U$ can be obtained. To assure the reliability of the samples fed to the subsequent classifier, the input samples to the classifier should have high confidence. There are two possible kinds of selection criteria, top $\eta_0$ potion of samples or general probability threshold. The distribution of membership in different clusters varies a lot. Our target is inner-cluster high confidence which can be viewed as a local property. Therefore, we choose the $\eta_0$ potion of samples with the highest confidence from each cluster, and increase $\Delta\eta$ per iteration to expand data capacity. $\eta_0$  is called initial selecting rate and $\Delta\eta$ is called increasing step.

\subsection{Local Boundary Cleaning and Augmentation}

Even though the samples selected by the fuzzy selecting strategy are supposed to have high confidence, there might still be some misassigned samples among the selected samples. These mistakes may produce severe destruction to the following classification and need be avoided.

The low dimensional embedded features for clustering are visualized with $t$-SNE \cite{van2008visualizing}. It is observed that most of the misassignments locate close to the boundary of the cluster. However, this phenomenon only exists with a single cluster, which is not obvious with multiple clusters. The different distributions of multiple clusters will affect each other during the process of $t$-SNE. Based on this observation, we start from single cluster and construct local boundary cleaning to restrict the influence of misassignments.

In visualization, misassignments may form an island independent of the correct samples. We first use density clustering to detect whether there are multiple clusters in the $t$-SNE projection of the $j^{\rm{th}}$ cluster $T_j$.  If true, remove the smaller clusters and the remainder is denoted as ${\hat T_j}$. The distribution of ${\hat T_j}$ is elliptical or circular because of the application of Euclidean distance in Eq. (1). The direction vectors ${\vec v_{\rm{L}}}$ and ${\vec v_{\rm{S}}}$ of the major axis and minor axis of the ellipse satisfy

\begin{align}
	{\tilde O_{i}}\left[ {{{\vec v}_{\rm{L}}},{{\vec v}_{\rm{S}}}} \right] = \left[ {\begin{array}{*{20}{c}}
			{{\lambda _{\rm{L}}}}&0\\
			0&{{\lambda _{\rm{S}}}}
	\end{array}} \right]\left[ {{{\vec v}_{\rm{L}}},{{\vec v}_{\rm{S}}}} \right],
\end{align}%
where ${\tilde O_{i}}$ is the two dimensional mapping of the selected samples in the $i^{\rm{th}}$ cluster, $\lambda _{\rm{L}}$  and $\lambda _{\rm{S}}$ are the corresponding characteristic values.
The major axis and minor axis are
\begin{align}
	{R_{\rm{L}}} = \max \left( {{{\tilde O}_{i}}{{\vec v}_{\rm{L}}}} \right),
	{R_{\rm{S}}} = \max \left( {{{\tilde O}_{i}}{{\vec v}_{\rm{S}}}} \right).
\end{align}%

Based on Eq. (5), the major axis can be obtained. The closer the samples are to the ellipse center, the more reliable the samples are. In order to weaken the negative impact of the misassignments, we only augment the high confident samples. Specifically, the ellipse is divided into 3 equal regions. The division scale on the major axis is ${{\sqrt 3 } \mathord{\left/{\vphantom {{\sqrt 3 } 3}} \right.\kern-\nulldelimiterspace} 3}:{{\sqrt 6 } \mathord{\left/{\vphantom {{\sqrt 6 } 3}} \right.\kern-\nulldelimiterspace} 3}:1$.  The ratio of the number of augmentations for the samples in the regions $a_1$, $a_2$ and $a_3$ is set as 5, 3, 1. Image augmentation techniques like random rotating and cropping are used to augment the samples within these regions. In this way, the number of reliable samples has been greatly increased and an augmented dataset $\hat X$ is generated to fine-tune the following classifier with the clustering labels as self-supervision.

\subsection{Self-supervised Classifier}

A classifier is incorporated to predict the target distribution automatically. With the fuzzy selecting strategy and local boundary cleaning, the data for augmentation are narrowed down. An augmented dataset can be generated by image augmentation techniques. Unlike the original dataset $X$, the augmented dataset $\hat X$ are labeled with the intermediate clustering result. These labels are taken as the supervision to fine-tune the classifier. In this sense, the classifier is trained in a self-supervised manner and no outside beforehand supervision is needed. Therefore, the whole package remains unsupervised learning.

The augmented data used to fine-tune the classifier are confidential samples, which enables the classifier to automatically learn reliable target distribution. The target distribution learnt by the classifier is more accurate and adaptive than that obtained by the handcrafted nonlinear mappings in previous deep clustering. In addition, the inputs to the classifier are raw images which contain more information than the embedded features used in most deep clustering. The information loss during the feature embedding process may distort the sample distribution, which may influence the obtained target distribution. Our solution can avoid such issue. After the fine-tuning the classifier with the augmented dataset, the target distribution of all the samples in the dataset $X$ for clustering can be predicted by the classifier as
\begin{align}
	{P_{\rm{target}}} = {f_{\rm{C}}}\left( {\rm{X}} \right),
\end{align}%
where ${P_{\rm{target}}}$ is the one-hot encoding label matrix. By this encoding method, the distribution of the samples belonging to the same class can be more compact. The objective function of Self-EvoC is also defined as a clustering loss evaluated by the Kullback-Leibler divergence as
\begin{align}
	{L_{\rm{C}}} = {\rm{KL}}(U\left| {{P_{\rm{target}}}} \right.) = \sum {U\log \frac{U}{{{P_{\rm{target}}}}}},
\end{align}%
where $U$ is the membership matrix and ${P_{\rm{target}}}$ is the resultant label matrix of the classifier. By minimizing the clustering loss in Eq. (7), the parameters of the feature extractor and the fuzzy clustering layer can be optimized.

\subsection{Optimization}
To optimize Self-EvoC, the update rule for the parameters can be obtained as
\begin{align}
	{\mu _j} = {\mu _j} - \frac{\alpha }{m_{\rm{b}}}\sum\limits_{i = 1}^{m_{\rm{b}}} {\frac{{\partial {L_{\rm{C}}}}}{{\partial {\mu _j}}}},
\end{align}%
and
\begin{align}
	w = w - \frac{\alpha }{m_{\rm{b}}}\sum\limits_{i = 1}^{m_{\rm{b}}} {\frac{{\partial {L_{\rm{C}}}}}{{\partial {z_i}}}} \frac{{\partial {z_i}}}{{\partial w}},
\end{align}%
where $\alpha$ is the learning rate and ${m_{\rm{b}}}$ is the number of samples within one training batch. If the change of the predicted classifier labels between two successive iterations is smaller than a threshold $\delta$, the training process will be stopped. The stopping criterion is formulated as
\begin{align}
1 - \frac{1}{N}\sum {P_{{\rm{target}}}^tP_{{\rm{target}}}^{t - 1}}  < \delta,
\end{align}%
where $P_{{\rm{target}}}^t$ and$P_{{\rm{target}}}^{t-1}$ are two successive target distribution in iteration $t$ and $t-1$. The detailed steps of the algorithm are summarized in Algorithm 1.

\begin{algorithm}[tb]
	\caption{Self-Evolutionary Clustering (Self-EvoC)}
	\label{alg:algorithm}
	\textbf{Input}: Dataset $X$, Local boundary cleaning mapping $\phi$, Fuzzifier value $m$, Initial selecting rate $\eta_0$, Increasing step $\Delta\eta$, Stopping threshold $\delta$, Maximum iteration number $N_{\rm{max}}$, Feature extractor $f_w$, Fuzzy clustering $f_{\rm{F}}$, Classifier$f_{\rm{C}}$ \\
	\textbf{Parameter}: $f_w$ parameters, $f_{\rm{F}}$ parameters, $f_{\rm{C}}$ parameters\\
	\textbf{Output}: Cluster assignment $P_{\rm{C}}$
	\begin{algorithmic}[1] 
		\STATE Initialize the clustering layer using Eqns (1), (2) and (3)
		\FOR{iter $\in{0,1,2,...,N_{\rm{max}}}$}
		\STATE Extract the features of the input samples ${\left\{ {{z_i} = {f_w}\left( {{x_i}} \right)} \right\}_{i = 1,2,...,n}}$.
		\STATE $U = {f_{\rm{F}}}\left( z \right)$.
		\STATE Select top $\eta_{\rm{iter}}$  samples from $X$ according to $U$ and get ${X_{{\eta _{\rm{iter}}}}}$.
		\STATE Generate augmented dataset $\hat X = \phi \left( {{X_{{\eta _{\rm{iter}}}}}} \right)$.
		\STATE Train classifier $f_{\rm{C}}$ with $\hat X$.
		\STATE Compute ${P_{\rm{target}}}$ by ${P_{\rm{target}}} = f_{\rm{C}}\left( X \right)$.
		\STATE Optimize $f_{\rm{F}}$ and $f_w$ with Eqns. (7), (8) and (9).
		\STATE ${\eta _{{\rm{iter}} + 1}} = {\eta _{{\rm{iter}}}} + \Delta \eta $.
		\IF { the Stopping criterion in Eq. (10) is met}
		\STATE Stop training.
		\ENDIF
		\ENDFOR
		\STATE Compute ${P_{\rm{C}}}$ by ${P_{\rm{C}}} = \arg \max \left( U \right)$.
		\STATE \textbf{return} ${P_{\rm{C}}}$
	\end{algorithmic}
\end{algorithm}
\begin{figure*}
	\begin{center}
		\footnotesize
		\begin{tabular}{c}
			\includegraphics[scale=0.77]{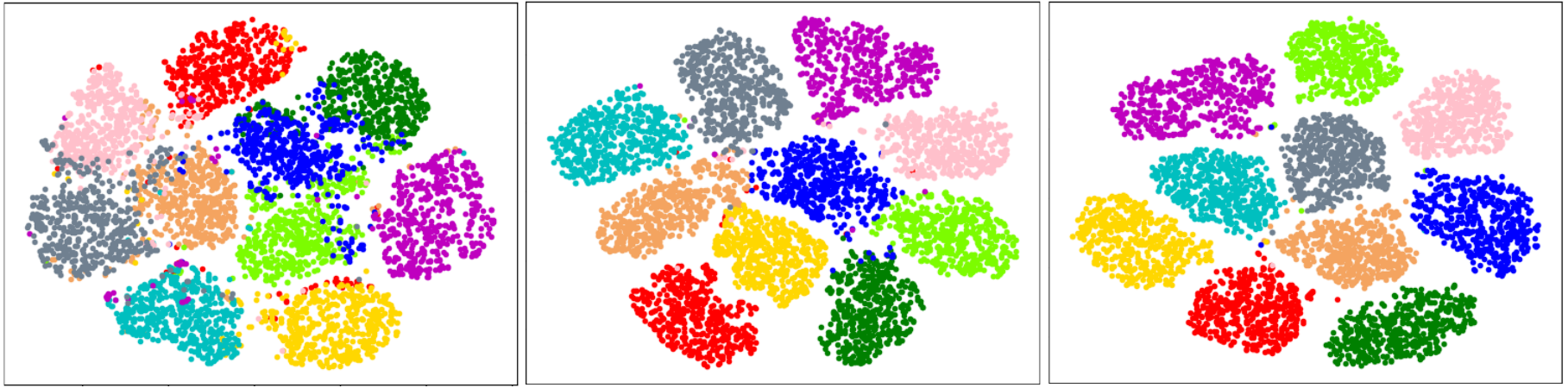}
		\end{tabular}
		\caption{\label{dcmg} {Visualization of clustering results on subset of MNIST during training.}}
	\end{center}
\end{figure*}

\section{Experiments}
\subsection{Datasets}
The proposed Self-EvoC framework is evaluated on three image datasets, including two handwritten digit datasets and one real world image dataset.

MNIST dataset consists of 60,000 handwritten digit images with the size of 28$ \times $28. We transform the image into a normalized 784 dimensional vector as input. 

USPS dataset contains 9,298 handwritten digit images with the size of 16$ \times $16 pixels. The image is transformed into a normalized 256 dimensional vector for clustering.

STL10 is a subset of the widely used benchmark ImageNet. It contains high dimensional real world images. There are 13,000 images in the dataset from 10 categories. All the images in STL10 are transformed into 96$ \times $96$ \times $3.

\subsection{Experiment Setup}
\textbf{Comparing methods. }We demonstrate the effectiveness of Self-EvoC framework by comparing it with the following representative clustering methods, $k$-means \cite{macqueen1967some}, FCM \cite{winkler2011fuzzy}, SEC \cite{nie2009spectral}, MBKM \cite{sculley2010web}, DEC \cite{xie2016unsupervised}, IDEC \cite{guo2017improved}, DFCM \cite{feng2020deep}, GrDNFCS \cite{feng2020deep}.
The results of the compared methods are from the literatures or obtained with open source code.

\textbf{Parameters setting. }Self-EvoC is a modular framework combining feature extractor, clustering and classifier to solve clustering problem. For different dataset, different feature extractors, clustering methods and classifiers can be selected. In our study, we choose the same fuzzy clustering layer implementation based on FCM for all the three datasets. The fuzzifier value is set as 1.4, 1.4 and 1.1 for MNIST, USPS and STL10 respectively. In our experiments, Autoencoder and off-the-shelf classifier are selected as the feature extractor and classifier for MNIST and USPS. Inspired by DEC, the structure of the Autoencoder is $d$-500-100-$L$-100-500-$d$, where $d$ is the input dimension and $L$=10 is the dimension of the embedded features. The classifier is set as a 4-layer convolutional neural network with 10 neurons as the output. We choose pre-trained model ResNet50 for feature extraction in the STL10 experiment. The classifier used for STL10 is SpinalNet \cite{kabir2020spinalnet} which is a powerful network imitating human brachial plexus. We choose SGD as the optimizer. The learning rate is 0.001 for all the three datasets. The stopping criterion threshold is set as $\delta$=0.001$\%$. The maximum iteration number is 20. All experiments are conducted on a server with Intel Xeon CPU Gold 6230 2.1GHz and GPU TITAN V. Tensorflow and Keras are used for implementation.

\textbf{Evaluation Metrics. }Clustering accuracy (ACC), normalized mutual information (NMI) and adjusted rand index (ARI) are used to evaluate the compared methods.
\begin{figure*}
	\begin{center}
		\footnotesize
		\begin{tabular}{c}
			\includegraphics[scale=0.78]{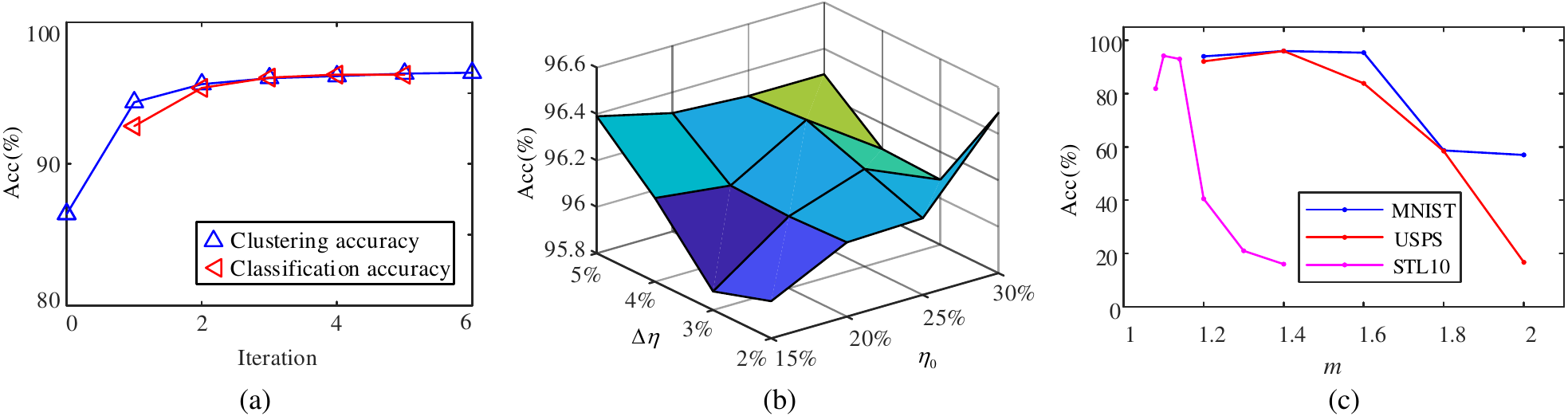}
		\end{tabular}
		\caption{\label{dcmg} {Performance evaluation of Self-EvoC. (a) Clustering accuracy and classification accuracy during training on MNIST. (b) Sensitivity of the initial selecting rate and the increaseing step on MNIST. (c) The variation of clustering accuracy with fuzzifier value on three datasets.}}
	\end{center}
\end{figure*}
\subsection{Results}
We report the comparison results on the 3 datasets in Table1. Self-EvoC has obtained the best performance among the compared methods. Deep clustering methods MBKM, DEC, IDEC, DFCM and GrDNFCS outperform traditional ones, which indicates the powerful feature extraction ability can improve the performance of the unsupervised clustering. With the help of fuzzy theory, the performance of GrDNFCS and Self-EvoC is better than other deep clustering methods. This suggests better features and target distribution can be generated by means of fuzzy modeling. The outperformance of Self-EvoC shows that the self-supervised classifier further improves the clustering performance.
\begin{table*}
	\centering
   	\begin{tabular}{lccccccccc}
		\toprule
		& & \textbf{MNIST} & & & \textbf{USPS} & & & \textbf{STL10} & \\
		& ACC & MNI & ARI & ACC & NMI & ARI & ACC & NMI & ARI \\
		\midrule
		$k$-means&0.5348 & 0.4999 & 0.3667 & 0.6679 & 0.6256 & 0.5450 & 0.7421 & 0.7234 & 0.5297\\
		FCM&0.5468 & 0.4816 & 0.3696 & 0.6634 & 0.6200 & 0.5393 & 0.7714 & 0.6813 & 0.5818\\		
		SEC&0.6273 & 0.6038 & 0.4859 & 0.6519 & 0.6488 & 0.4936 & 0.6535 & 0.5601 & 0.4584\\
		MBKM&0.5443 & 0.4482 & 0.3685 & 0.6287 & 0.5993 & 0.5105 & 0.6894 & 0.6794 & 0.5275\\	
		DEC&0.8653 & 0.8369 & 0.8029 & 0.7278 & 0.7352 & 0.6622 & 0.8944 & 0.8201 & 0.7826\\
		IDEC&0.8801 & 0.8638 & 0.8325 & 0.7513 & 0.7595 & 0.6791 & 0.9005 & 0.8285 & 0.7974\\	
		DFCM&0.8817 & 0.8654 & 0.8337 & 0.7536 & 0.7636 & 0.6815 & 0.9020 & 0.8296 & 0.8009\\	
		GrDNFCS&0.9145 & 0.9074 & 0.8626 & 0.7652 & 0.7761 & 0.6903 & 0.9143 & 0.8410 & 0.8252\\	
		Self-EvoC&\textbf{0.9638} & \textbf{0.9209} & \textbf{0.9236} & \textbf{0.9747} & \textbf{0.9336} & \textbf{0.9504} & \textbf{0.9420} & \textbf{0.8808} & \textbf{0.8747}\\	
		\bottomrule
	\end{tabular}
 \caption{Clustering Performance of Different Algorithms in Terms of ACC, NMI and ARI}
\label{tab:booktabs}
\end{table*}

We further explore the variation of the sample distribution by visualizing the embedded space during the clustering process. The   visualization on a random selected subset with 1,000 samples in each cluster in MNIST is shown in Figure 3. From left to right, the three images correspond to the visualization of the initial stage, the middle stage and the final result of the training process. It could be seen that along with the training progress, the gap between two clusters gradually increase and the clusters become highly compact. According to \cite{xie2016unsupervised} and the initial clustering results in our experiments, digits 4 and 9 colored by light green and blue are easily to be confused. However, classifier-based target distribution takes the advantage of the powerful feature representation capability of network model and can separate these two digits clearly. These results show the superiority of Self-EvoC.

To reveal how the classifier affects the performance of Self-EvoC, the accuracy curves of the clustering and the classifier are shown in Figure 4(a). The curves are obtained from MNIST. In the MNIST experiment, the training will stop at the $6^{\rm{th}}$ iteration which means the classifier predicts the target distribution for 5 times. The $t^{\rm{th}}$ classifier prediction is based on the $t-1^{\rm{th}}$ cluster assignment. It could be seen that the prediction accuracy of the classifier is higher than the clustering counterpart, which explains the evolved performance of Self-EvoC. As a result, the feature extractor and the fuzzy clustering centers can be further optimized.

The sensitivity of the important hyper-parameters are analyzed including the initial selecting rate $\eta_0$, increasing step $\Delta\eta$ and fuzzifier value $m$. Figure 4(b) gives the clustering accuracy under different setting of $\eta_0$ and $\Delta\eta$. It can be seen that Self-EvoC performs slightly better with the increasing of these two parameters. However, this doesn’t mean the higher the better. Oversized $\eta_0$ and $\Delta\eta$ will lead to misassigned samples and deteriorate the performance. Figure 4(c) shows the variation of the clustering accuracy with respect to the fuzzifier value. It can be observed that with the increase of fuzzfizer value, the accuracy first increases and then decreases. The best fuzzfizer value for STL10 is 1.1 and the best fuzzifier value for MNIST and USPS is 1.4. The more complex the dataset, the smaller the fuzzifier value is expected.

\section{Conclusion}
This paper constructs a modular Self-EvoC framework, which combines feature extractor, clustering and classifier. Off-the-shelf classification algorithm is used to train a target distribution generation model, which can generate more reliable and adaptive target distribution as compared with the previous handcrafted nonlinear mappings. Fuzzy selecting strategy and local boundary cleaning are designed to select the most confidential samples within the intermediate obtained clusters. The selected data are augmented and used for fine-tuning of the classifier to generate better target distribution. The intermediate generated cluster labels are used as the supervision. In this way, the classifier is trained in a self-supervised manner. The unsupervised clustering is enhanced by self-supervised classifier. The Self-EvoC gets self-evolved. Extensive experiments on three benchmarks demonstrate the superior performance of Self-EvoC. This work has shed some new light into deep clustering.

\section*{Acknowledgments}
This work is supported by National Key R$\&$D Program of China 2018AAA0101501, Science and Technology Project of SGCC (State Grid Corporation of China): Fundamental Theory of Human-in-the-loop Hybrid-Augmented Intelligence for Power Grid Dispatch and Control.

\bibliographystyle{named}
\bibliography{ijcai22}

\end{document}